\documentclass[10pt,twocolumn,letterpaper]{article}

\usepackage{cvpr}              

\usepackage[dvipsnames,table]{xcolor}

\usepackage{cuted}
\usepackage{afterpage}

\newcommand{\metrics}{PSNR$\uparrow$ & SSIM$\uparrow$ & LPIPS$\downarrow$}
\newcommand{\hmerge}[1]{\multicolumn{3}{c|}{#1}}
\newcommand{\hmergl}[1]{\multicolumn{3}{c}{#1}}

\newcommand{\lmetrics}{Avg. & $90^{\mathrm{th}}$ & $95^{\mathrm{th}}$ & $99^{\mathrm{th}}$}
\newcommand{\lmerge}[1]{\multicolumn{4}{c|}{#1}}
\newcommand{\lmergl}[1]{\multicolumn{4}{c}{#1}}

\definecolor{tabf}{rgb}{1, 0.7, 0.7}   
\definecolor{tabs}{rgb}{1, 0.85, 0.7}  
\definecolor{tabt}{rgb}{1, 1, 0.7}     
\newcommand{\cccf}[1]{\cellcolor{tabf}{\textbf{#1}}}
\newcommand{\cccs}[1]{\cellcolor{tabs}{#1}}
\newcommand{\ccct}[1]{\cellcolor{tabt}{#1}}

\definecolor{cvprblue}{rgb}{0.21,0.49,0.74}
\usepackage[pagebackref,breaklinks,colorlinks,citecolor=cvprblue]{hyperref}
\usepackage{marvosym}

\def\paperID{173} 
\def\confName{3DV\xspace}
\def\confYear{2025\xspace}

\title{Para-Lane: Multi-Lane Dataset Registering Parallel Scans \\ for Benchmarking Novel View Synthesis}
\author{Ziqian Ni$^1$ \quad Sicong Du$^1$ \quad Zhenghua Hou$^1$ \quad Chenming Wu$^2$ \quad Sheng Yang$^{1}$\textsuperscript{\Letter} \\
{\normalsize $^1$Autonomous Driving Lab, CaiNiao Inc., Alibaba Group \quad
$^2$Baidu Research}\\
{\tt\small \{niziqian.nzq,dusicong.dsc,houzhenghua.houzhe,shengyang\}@alibaba-inc.com}
}

\begin{document}
\maketitle
\let\thefootnote\relax\footnote{\Letter: Corresponding author.}

\begin{strip}
\begin{minipage}{\textwidth}\centering
\vspace{-30pt}
\includegraphics[width=\textwidth]{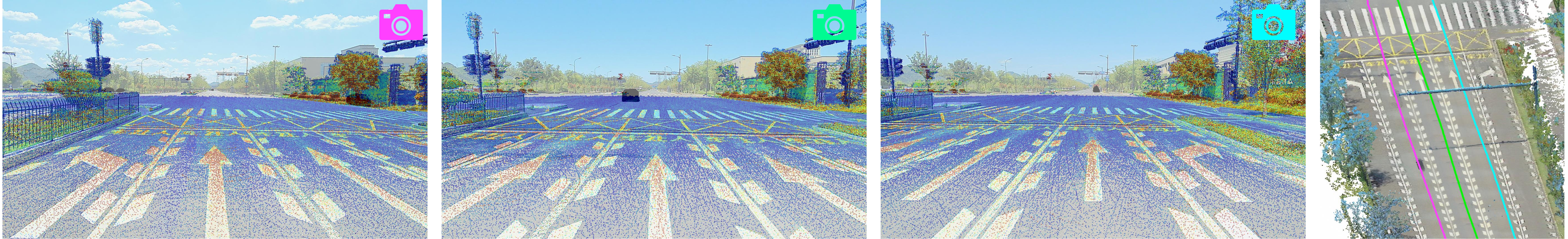}
\captionof{figure}{Our work introduces the first real-world multi-lane dataset for evaluating the novel view synthesis capabilities of recent reconstruction approaches for autonomous driving. Public urban roads are scanned using multi-pass trajectories with three laser scanners, a front-view camera, and four surround-view cameras. Frame-wise poses are accurately aligned through LiDAR mapping and multi-modal Structure-from-Motion techniques. Here, we present example images captured from close positions in three aligned cross-lane sequences, with a shared point cloud projected onto the images based on our optimized camera-LiDAR poses.}
\label{fig:teaser}
\end{minipage}
\end{strip}

\begin{abstract}
To evaluate end-to-end autonomous driving systems, a simulation environment based on Novel View Synthesis (NVS) techniques is essential, which synthesizes photo-realistic images and point clouds from previously recorded sequences under new vehicle poses, particularly in cross-lane scenarios. Therefore, the development of a multi-lane dataset and benchmark is necessary. While recent synthetic scene-based NVS datasets have been prepared for cross-lane benchmarking, they still lack the realism of captured images and point clouds. To further assess the performance of existing methods based on NeRF and 3DGS, we present the first multi-lane dataset registering parallel scans specifically for novel driving view synthesis dataset derived from real-world scans, comprising 25 groups of associated sequences, including 16,000 front-view images, 64,000 surround-view images, and 16,000 LiDAR frames. All frames are labeled to differentiate moving objects from static elements. Using this dataset, we evaluate the performance of existing approaches in various testing scenarios at different lanes and distances. Additionally, our method provides the solution for solving and assessing the quality of multi-sensor poses for multi-modal data alignment for curating such a dataset in real-world. We plan to continually add new sequences to test the generalization of existing methods across different scenarios. The dataset is released publicly at the project page: \href{https://nizqleo.github.io/paralane-dataset/}{https://nizqleo.github.io/paralane-dataset/}.
\end{abstract}
\section{Introduction}
\label{sec:intro}

As a widely investigated technique in 3D vision, novel view synthesis (NVS)~\cite{mildenhall2021nerf,Kerbl20233dgs} is utilized in two primary ways in the development of autonomous driving systems: (1) It facilitates the transfer of data for trained perception or end-to-end models across different vehicle products and sensor configurations~\cite{chen2023e2esurvey}; (2) It generates sensor frames with realistic geometry and appearance from various viewpoints for closed-loop simulations, particularly in sensor-to-control scenarios~\cite{yang2024drivearena}.

However, most existing NVS methods in autonomous driving, such as ~\cite{Kerbl20233dgs,yan2024street}, primarily focus on evaluating novel views based on interpolation quality rather than lateral viewpoint shifts, i.e., cross-lane NVS. This is due to the lack of datasets and benchmarks specifically designed for this purpose. As a result, the full potential impact of these new algorithms has not been fully demonstrated. Consequently, current simulation platforms have primarily validated the effectiveness of testing strategic changes in longitudinal (speed) behavior and motion planning, while evaluations of lateral (path) planning remain less convincing.

Unfortunately, collecting multi-lane ground truth data within a real-world scene is intrinsically difficult. As a result, XLD~\cite{li2024xld} chooses to use a simulation platform, Carla~\cite{Dosovitskiy17}, to render synthetic data with perfect parameters such as intrinsic, extrinsics, rolling-shutter appearances, and sensor frame poses. They have two primary limitations for further use: First, creating synthetic scenes with high-definition mesh models and materials requires meticulous adjustments for artistic and coherent shading. This process is expensive and must be completed before efficiently generating photo-realistic frames for evaluating driving failure cases on-board. Second, there are still challenging photo-realistic issues, such as inherent noise of real-sensors, and fog-wind-fluid caused dynamics, causing artifacts and domain transfer costs~\cite{zhang2024resimadzeroshot3ddomain}. Therefore, even though sensor data obtained from real recordings may have issues with imprecise parameters, they are an indispensable part of evaluating cross-lane NVS quality.
If we aim to collect the data in a single pass, then a super large structure to rigidly mount multiple cameras is required, however, a typical width of a round lane is about 3-4 meters, which is intricate to design and manufacture such a structure with stable dynamics. To tackle this difficulty, our method opts for multi-pass data collection.

This paper focuses on addressing the challenges of creating a real-world dataset using a multi-pass collection scheme for cross-lane NVS benchmarks by tackling the following issues. First, commonly used inertial navigation systems (INS) for obtaining pseudo ground-truth vehicle trajectories in most datasets~\cite{geiger2012kitti} are insufficient for aligning temporally adjacent sensor frames. This is because the data association between consecutive frames of exterceptive sensors does not directly participate in the maximum-a-posterior pose estimation. Instead, {temporal} alignment is achieved through dual filtering of the RTK-IMU trajectory, and {then composite with extrinsic parameters between IMU and exteroceptive sensors calibrated during end-of-line}.
Second, {pixel-to-point} mapping from camera frames to LiDAR frames is imprecise if we rely solely on the trajectory and LiDAR-IMU-cameras extrinsics after pose estimation{, which becomes worse when we need to remap across multiple scans.} Without establishing and resolving cross-modal feature correspondences between exteroceptive sensors~\cite{pascoe2015robust,song2016robust}, the mapping lacks accuracy. {We have addressed these two issues through our two-phase pose optimization (Sec.~\ref{sec:dataset:slam}).}

We develop a unified framework to construct the Para-Lane dataset, featuring a two-phase pose optimization mechanism for aligning data from exteroceptive sensors both temporally and spatially. Additionally, we implemented an autonomous system equipped with LiDAR and camera sensors to capture data, which is then processed using our proposed framework for cross-modal alignment.
As the first real-world dataset for cross-lane scenarios, we benchmark mainstream methods for driving NVS based on either NeRF or 3DGS. Our findings could inspire further research and enhance end-to-end driving simulations, and the dataset will be released publicly, ultimately and hopefully accelerating the research and development of autonomous driving products.
In summary, our work offers three key contributions:

\begin{itemize}
\item We curate the first real-world cross-lane dataset, dubbed Para-Lane, for evaluating NVS capabilities. It includes ample LiDAR, front-view, and surround-view camera data. Our dataset ensures that all sequences are annotated and grouped for easier benchmarking.
\item We propose a two-stage framework to precisely align exteroceptive sensors using cross-modal correspondences, demonstrating effectiveness in alignment metrics.
\item We evaluate recent NeRF and 3DGS methods, including those designed for autonomous driving scenes, on our curated dataset, offering insights into NVS performance with lateral viewpoint shifts.

\end{itemize}

\section{Related Work}
\label{sec:rel_work}

\subsection{Autonomous Driving Datasets for NVS} 
In autonomous driving, there exists a number of available datasets, such as KITTI~\cite{geiger2012kitti}, KITTI-360~\cite{liao2022kitti360}, CityScapes~\cite{cordts2016cityscapes}, Waymo Open Dataset~\cite{sun2020waymo}, nuScenes~\cite{caesar2020nuscenes}, LiDAR-CS~\cite{fang2024lidar} and WayveScenes101~\cite{zürn2024wayvescenes101datasetbenchmarknovel}. In decades, they are regarded as the foundation of numerous autonomous driving solutions and algorithms~\cite{hu2023uniad,liao2023MapTR,chen2023e2esurvey}. However, the community lacks datasets that are specifically or compatible to evaluate NVS tasks, due to the booming requirements of end-to-end autonomous driving research. 
A very recent work, Open MARS Dataset~\cite{li2024openmars}, also features multiple laser scanners and cameras for driving scenes. However, their focus is performing collaboratively multi-agent and multi-traversal data collection, with an emphasis on obtaining spatially nearby sequences, instead of multi-lane sequences. Moreover, the multi-pass frame registration algorithm is not disclosed in their work.
The XLD~\cite{li2024xld} dataset, introduced earlier this year, serves as a reliable resource based on synthetic scenes and rendering. However, as discussed in Sec.~\ref{sec:intro}, we argue that real-world datasets are more essential and reliable for comprehensive evaluation, though there are many challenges to curate such a dataset.

\subsection{Multi-Sensor Data Alignment} 
We categorize the multi-sensor dataset alignment into single-modal (camera or LiDAR) and multi-modal (camera and LiDAR) alignments, both of which are crucial for creating a high-quality real-world dataset for NVS evaluation.


Ensuring precise alignment of data from multiple sensors is essential. While a unified multi-sensor SLAM framework can achieve this, the industrial community typically handles these steps separately for large-scale scene production. For Level-4 unmanned vehicles, during High-definition Map (HDMap) production, LiDAR mapping has been a well-established process~\cite{yang2018pgo} and is now standard for mass production and vehicle deployment. For cameras, Structure-from-Motion (SfM)~\cite{snavely2006photo,scho2016sfm} and multi-view stereo~\cite{moulon2016openmvg,Xu2020ACMP} are commonly used for solving frame poses and performing dense geometric reconstruction.

To address the cross-modal data association issue—linking cameras and LiDAR—we need to establish explicit correspondences between LiDAR points and camera pixels for a densely coupled pose estimation. There are two categories of methods based on their inputs: the first category operates between LiDAR frames and camera frames~\cite{jing2022dxqnet}, which is inefficient for generating large sequences. Therefore, assuming LiDAR mapping provides a sufficiently accurate trajectory, the second category—our chosen approach—works between LiDAR sequences and camera frames~\cite{zhou2014colormapopt,li2024vxp}. While some methods primarily use ray-casted depth information, we found that incorporating the intensity channel from LiDAR measurements is beneficial. This allows for the use of both sparse and dense photometric loss alongside geometric loss, as demonstrated in various RGB-D reconstruction methods~\cite{Whelan15rss,dai2017bundlefusion}. To establish photometric loss, we identified Normalized Information Distance (NID)~\cite{ming2004similarity,pascoe2015robust} as the most effective metric for linking these two types of channels, which can assess the quality of multi-modal registration.

\subsection{NeRF and 3DGS Approaches for NVS} 
Beyond the classical dense reconstruction methods that utilize Truncated Signed Distance Function (TSDF)~\cite{dai2017bundlefusion} and Surfels~\cite{Whelan15rss} to represent large-scale scenes. NeRF~\cite{mildenhall2021nerf} and 3DGS~\cite{Kerbl20233dgs} methods are profound innovations in the enhancement of representing geometric and appearance details, respectively. 
For instance, Block-NeRF~\cite{tancik2022block} is a pioneering work that addresses the reconstruction of large-scale urban scenes through division.
MARS~\cite{wu2023mars} is a modular, instance-aware simulator built on NeRF, which separately models dynamic foreground instances and static background environments. 
UniSim~\cite{yang2023unisim} converts recorded logs into realistic closed-loop multi-sensor simulations, incorporating dynamic object priors and using a convolutional network to address unseen regions. Rather than implicitly representing scenes (NeRF) that lack of flexibility of editing and labeling those reconstructed assets, explicitly representing scenes (3DGS) has aroused the interest of many industrial autonomous driving teams~\cite{yan2024street} to operate on their domain-specific database. 
Our dataset and benchmark scanned from the real world are specifically designed to evaluate the performance of cross-lane NVS methods. Some of these methods~\cite{Kerbl20233dgs,chen2023pvg,cheng2024gaussianpro,Lu2024scaffoldgs,Huang20242DGS,yan2024street}, considering the availability of their code, are used as baselines in our evaluation. 

\section{Curation of Para-Lane Dataset}
\label{sec:dataset}

\subsection{Sensor Setup}

To collect and process raw data for our Para-Lane dataset, we implemented an autonomous system equipped with various sensors for street operation. Specifically, to meet the input data specifications for NVS approaches, we installed one front-view camera with a $90^\circ$ Field-of-View (FOV) capturing $1920 \times 1080$ images at 10Hz, along with four surround-view cameras featuring $190^\circ$ fisheye lenses capturing $1920 \times 1080$ images at 10Hz. Additionally, we used three 3D laser scanners with 32 LiDAR channels from $+15^\circ$ to $-55^\circ$ on the vertical FOV capturing at 10Hz. All frame timestamps are synchronized at the hardware level, and we combine points from the three laser scanners into a single LiDAR frame after motion compensation.

\begin{figure}[t]
\includegraphics[width=\linewidth]{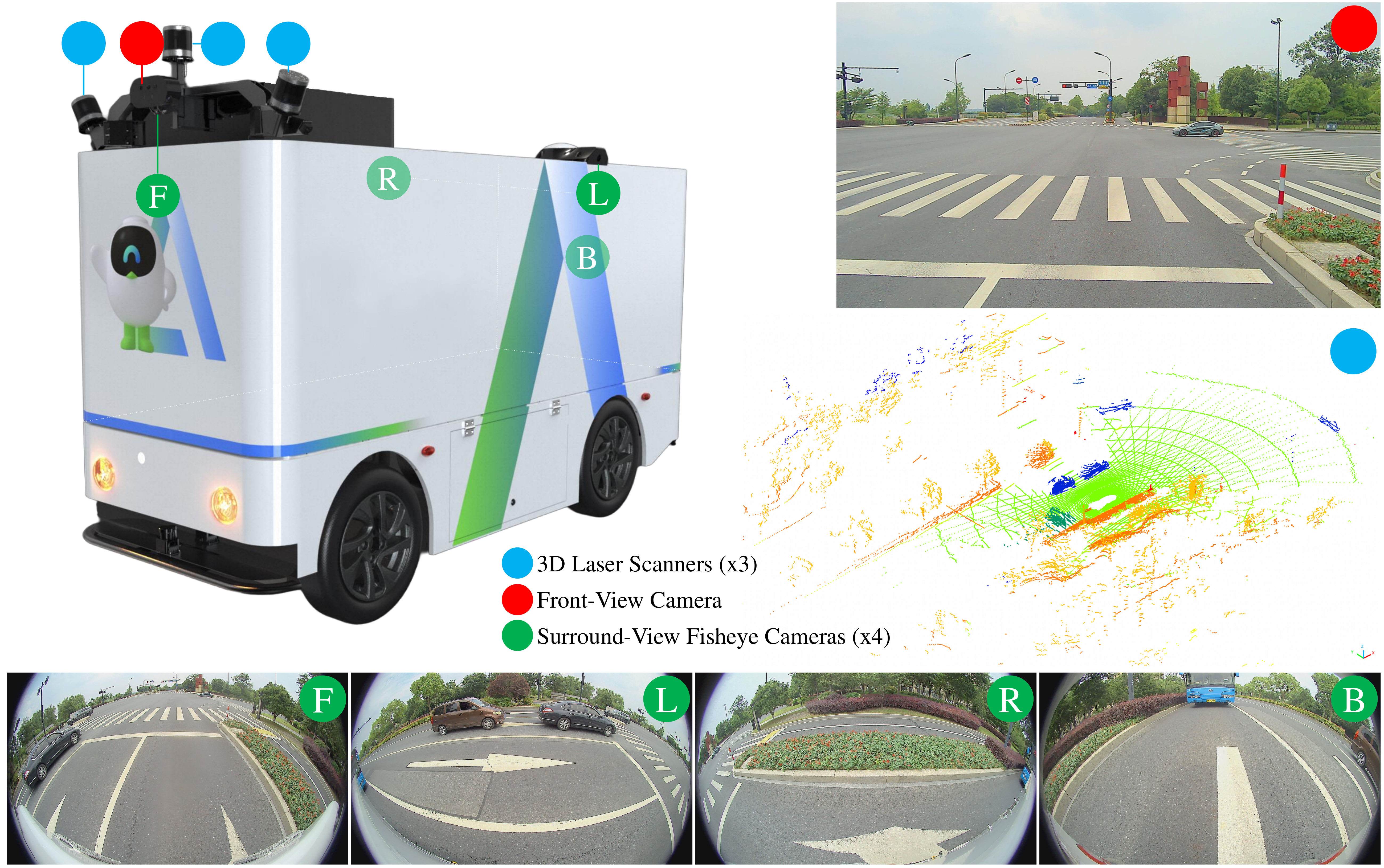}
\caption{Sensor assembly and sample frames of our data collection unmanned vehicle, the right fisheye camera is mounted symmetrically opposite to the left fisheye, and the back fisheye is located at the center of the back-side.}
\label{fig:assemble}
\end{figure}

Fig.~\ref{fig:assemble} illustrates our autonomous hardware and its assembly, and we refer readers to our {accompanying} calibration parameters {in the dataset} for more details. Besides the sensors depicted in the figure, we have additional sensors inside, such as an Inertial Navigation System (INS) for obtaining a high-quality initial trajectory prior to data alignment. In our dataset, we provide all necessary relative transformations between {grouped} sequences and a reference coordinate system.

\subsection{Multi-pass Data Acquisition and Processing}

We selected clear sunny days with uncongested road conditions to drive through each parallel lane in the same direction. The data was collected in an anonymous city (details will be disclosed after the review period), consisting of 75 sequences grouped into 25 scenes. Each scene includes three sequences from different lanes, sharing the same start and end positions orthogonal to the road direction, covering approximately 150 meters. As our collection vehicle traveled on public roads, its speed fluctuated, resulting in sequence durations ranging from 10 to 45 seconds, with a median of 20 seconds.

After data collection, we anonymized the images by obscuring vehicle license plates and pedestrian faces. We employed SAM~\cite{Kirillov2023sam}, followed by a manual quality inspection, to segment dynamic elements from static foregrounds and backgrounds, ensuring the dataset is free of ethical issues.

\subsection{Two-phase Pose Optimization}
\label{sec:dataset:slam}

\noindent \textbf{Phase 1: LiDAR mapping.} 
Given the {initial trajectory} from the RTK/INS sensor, we apply~\cite{yang2018pgo} to construct a LiDAR map $\mathcal{L}^\mathbb{G}$ in a reference coordinate $\mathbb{G}$, which involves two components: odometry and loop refinement. We focus on the first component—offline LiDAR odometry—by implementing an enhanced offline LIO system~\cite{shan2020liosam} that utilizes both LOAM features~\cite{zhang2014loam} and dense scan-to-submap point cloud registration~\cite{segal2009gicp} for robustness across various mapping scenarios. For the second component, loop closure and pose graph optimization, we begin with the coarse matching of submaps using Predator~\cite{huang2021predator} and SuperLine3D~\cite{zhao2022superline3d}. We then evaluate the matching scores to select candidate submap pairs for fine registration~\cite{segal2009gicp}, establishing precise relative constraints for the pose graph optimization problem. Relative poses between sequences are determined through multi-pass loop closure associations and joint optimization.

\begin{figure}[t]
\includegraphics[width=\linewidth]{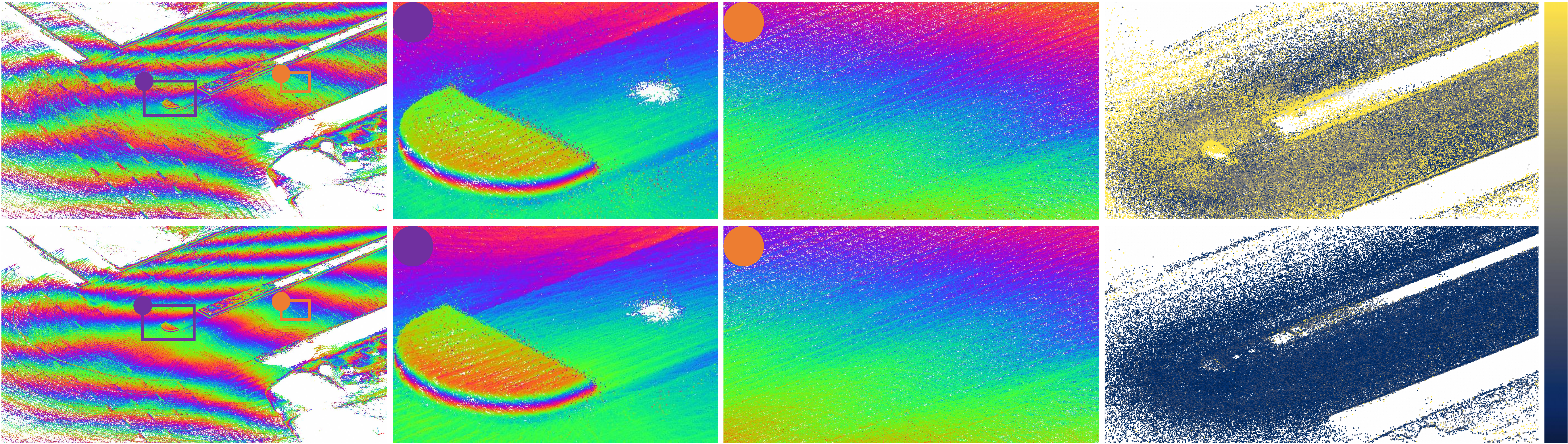}
\caption{LiDAR map stitching quality visualized in both 20cm periodical height ramp {in rainbow} (left columns) and 10cm cividis colormap reflecting distance with their reconstructed mesh (the right column). Both the error map and zoomed-in views reflect that these refined LiDAR frame poses (the second row), compared to the initial RTK trajectory (the first row), have achieved a thinner stitched cloud with fewer hovering noisy points due to better frame poses.}
\label{fig:lidarmap}
\end{figure}

To verify the quality of LiDAR mapping, we use the thickness of structural objects as a reference metric. We first perform mesh reconstruction using VDBFusion~\cite{vizzo2022sensors} and Marching Cubes~\cite{lorensen87marchingcube} with the solved LiDAR frame poses (both voxel size and truncation distance set to 5 cm) to create a triangle mesh representing the maximum-a-posteriori locations of watertight object surfaces. We then calculate the Mean Absolute Error (MAE) and Root Mean Square Error (RMSE)~\cite{millane2018cblox} by comparing the mesh with the stitched point cloud to assess thickness. Tab.~\ref{tab:lidarmapping} and Fig.~\ref{fig:lidarmap} demonstrate the effectiveness of {achieving a thinner LiDAR map after the combination of multiple scaned sequences.}

\noindent \textbf{Phase 2: Registering images to the LiDAR map.} 
After obtaining the LiDAR map $\mathcal{L}^\mathbb{G}$ and the pose of each LiDAR frame, we register the images $\bigcup_i \{\mathbf{C}_i\}$ captured by our camera sensors to $\mathcal{L}^\mathbb{G}$. This involves determining the pose of each camera frame $\mathbf{C}_i$ relative to the reference coordinate, denoted as $\mathbf{T}_{\mathbb{C}_i}^\mathbb{G}$. We begin by coarsely initializing these poses using linear-slerp interpolation~\cite{buss2001linearslerp} between the two adjacent LiDAR frames and an extrinsic parameter.

We then refine the coarse camera poses by formulating the following factor graph optimization problem~\cite{grisetti2010tutorial} to optimize all camera poses $\mathcal{C}^\mathbb{G} \triangleq \bigcup_i \{\mathbf{T}_{\mathbb{C}_i}^\mathbb{G}\}$ and the 3D positions of visual landmarks $\mathcal{P}^\mathbb{G} \triangleq \bigcup_j \{\mathbf{p}_j^{\mathbb{G}}\}$ using an expectation-maximization (EM) strategy~\cite{bowman2017em}:
\begin{equation}
\min_{\mathcal{C}^\mathbb{G}, \mathcal{P}^\mathbb{G}} 
  {\mathbf{E}^1(\mathbf{C}_i, \mathbf{p}_j}) +
  {\mathbf{E}^2(\mathbf{p}_j, \mathcal{L}^\mathbb{G}}) +
  {\mathbf{E}^3(\mathbf{C}_i, \mathbf{p}_j}, \mathbf{L}_i),
\label{equ:sfm}
\end{equation}
where $\mathbf{E}(a, b[, c]) = \sum -\log(\mathbf{f}_{a, b[, c]})$ is the sum of negative log-likelihood of a type of valid factor constraints $\mathbf{f}$, making their scale factors become irrelevant constants. $\mathbf{L}_i \triangleq \pi((\mathbf{T}_{\mathbb{C}_i}^\mathbb{G})^{-1} \cdot \mathcal{L}^\mathbb{G})$ is the intensity image ray-casted from $\mathcal{L}^{\mathbb{G}}$ at $\mathbf{T}_{\mathbb{C}_i}^\mathbb{G}$. 
Specifically, these factors $\{\mathbf{f}\}$ are designed as:

\begin{table}[t]

\caption{Quantitative metrics for LiDAR mapping. We choose to sample and evaluate the MAE and RMSE of stitched LiDAR frames (in centimeters).}\vspace{-4pt}
\centering
\fontsize{8pt}{9.6pt}\selectfont
\setlength{\tabcolsep}{4pt}
\begin{tabular}{l|cccc|cccc}
\toprule
       &  \lmerge{MAE $\downarrow$}            &  \lmergl{RMSE $\downarrow$} \\
Metrics&  \lmetrics               &  \lmetrics  \\
\midrule 
Before & 4.5 & 5.8 & 5.9 & 7.3 & 7.4 & 9.8 &10.3 &12.6  \\
After  & \cccf{1.3} & \cccf{1.4} & \cccf{1.4} & \cccf{1.7} & \cccf{2.9} & \cccf{3.6} & \cccf{4.1} & \cccf{7.1}  \\
\bottomrule
\end{tabular}
\label{tab:lidarmapping}
\end{table}

\textbf{$\mathbf{f}^1$: Structure-from-motion (SfM) for cameras.} We use SuperGlue~\cite{wang2019superglue} to associate multiple camera frames with joint observations on sparse features.
We formulate this factor using the reprojection error~\cite{scho2016sfm}, which quantifies the error by projecting the 3D positions of landmarks $\mathbf{p}_j^\mathbb{G}$ onto the image plane of the corresponding frame pose $\mathbf{T}_{\mathbb{C}_i}^\mathbb{G}$:
\begin{equation}
  \mathbf{f}_{\mathbf{C}_i, \mathbf{p}_j}^{1} \propto \exp(-\frac{1}{2} \|
  \hat{v}_j - \mathbf{K} ( (\mathbf{T}_{\mathbb{C}_i}^\mathbb{G})^{-1} \cdot \mathbf{p}_j^\mathbb{G} ) 
\|_{\Omega_{1}}^2),
\label{equ:sfm1}
\end{equation}
where $\| \mathbf{e} \|_\Omega^2 \triangleq \mathbf{e}^\top \Omega^{-1} \mathbf{e}$ represents the squared Mahalanobis distance with the covariance matrix $\Omega$. Here, $\hat{v}_j$ is the corresponding pixel observed in $\mathbf{C}_i$, and $\mathbf{K}(\cdot)$ denotes the perspective projection and rectification function with respect to the intrinsic and distortion parameters.

\textbf{$\mathbf{f}^2$: SfM points and LiDAR map registration factors.} 
Longitudinal driving sequences often lack sufficient parallax to accurately estimate depth for the sparse landmarks $\mathcal{P}^\mathbb{G}$. Therefore, we constrain the absolute positions of these landmarks once they are effectively associated with nearby LiDAR points. The unary prior factor for each landmark is defined as:
\begin{equation}
  \mathbf{f}_{\mathbf{p}_j, \mathcal{L}^\mathbb{G}}^{2} \propto \exp(-\frac{1}{2} \|
  \mathbf{p}_j^\mathbb{G} - \Psi(\mathcal{L}^\mathbb{G}, \mathbf{p}_j^\mathbb{G})
\|_{\Omega_{2}}^2).
\label{equ:sfm2}
\end{equation}

We found that the choice between point-to-point and point-to-plane formulations is not critically important; however, the point-picking strategy $\Psi(\cdot)$ is significant. Specifically, we select a group of candidate 3D LiDAR points based on the current nearest search for $\mathbf{p}_j^\mathbb{G}$, and calculate the tangential distance between these points and the rays emitted from multiple camera observations $\mathcal{R}(\mathbb{C}_i, \hat{v}_j^{\mathbb{C}_i})$ as a joint weight for the final prior position.

\textbf{$\mathbf{f}^3$: Cross-modal sparse and dense factors.} Constructing linkage among sparse 3D points through the previous factor $\mathbf{f}^2$, in our scenario, is not sufficient enough for a camera-pixel to LiDAR-point level data integration. Inspired by the tightly-coupled multi-modal registration used in RGB-D reconstruction~\cite{dai2017bundlefusion}, we use a combined sparse-and-dense factor, with SuperGlue~\cite{wang2019superglue} again for providing sparse constraints, and we define cross-modal photometric error~\cite{stein2011densepholoss} for dense constraints as:
\begin{align}
  \mathbf{f}_{\mathbf{C}_i, \mathbf{p}_j, \mathbf{L}_i}^{3} \propto \exp(
  &-\frac{1}{2} \| \hat{w}_j - \mathbf{K} ((\mathbf{T}_{\mathbb{C}_i}^\mathbb{G})^{-1} \cdot \mathbf{p}_j^\mathbb{G} ) \|_{\Omega_3^s}^2 \\
  &-\frac{1}{2} \|
  \mathbf{C}_i \ominus \mathbf{L}_i \|_{\Omega_3^d}^2 ),
\label{equ:sfm3}
\end{align}
where the first half focuses on maintaining sparse relationships between input and ray-casted frames, while the second half addresses dense matching. To establish sparse correspondences between ray-casted feature points $\hat{w}_j$ and SfM points $\mathbf{p}_j^\mathbb{G}$, we utilize feature points extracted and matched in $\mathbf{C}_i$ to connect these 2D-3D relationships. The operator $\ominus$ represents $E(\xi)$ as defined in Equation 11 of~\cite{stein2011densepholoss}, which employs image gradients to slightly adjust the camera pose.

{\textbf{Parameters.}
We set the number of Superglue correspondences to 500 for both $\mathbf{f}^1$ and $\mathbf{f}^3$. We use parameters $\Omega_1 = 1$, $\Omega_2 = 0.3 \cdot \mathbf{I}_3$, $\Omega_3^s = 1$, and $\Omega_3^d = 0.01$ to perform the maximization step using Levenberg-Marquardt (LM) with 50 iterations. During optimization, we recalculate the point picking strategy $\Psi(\cdot)$ in $\mathbf{f}^2$ and the ray-casted image $\mathbf{L}_i$ from $\pi(\cdot)$ in $\mathbf{f}^2$ once every 3 iterations.}

\textbf{Evaluation and Ablation Study.} To evaluate the quality of multi-camera alignment, we utilize the Normalized Information Distance (NID) proposed in Equation 3 of~\cite{pascoe2015robust} as a quantitative metric. For each pair of original grayscale images and LiDAR intensity images rendered from a given pose, we compute an initial NID by discretizing the continuous pixel values into 64 bins. We then slightly shift the original image along the UV coordinates to identify a position that yields a local minimum of the NID. The offset used to find this local minimum is defined as NID-Loss.

\begin{figure}[t]
\includegraphics[width=\linewidth]{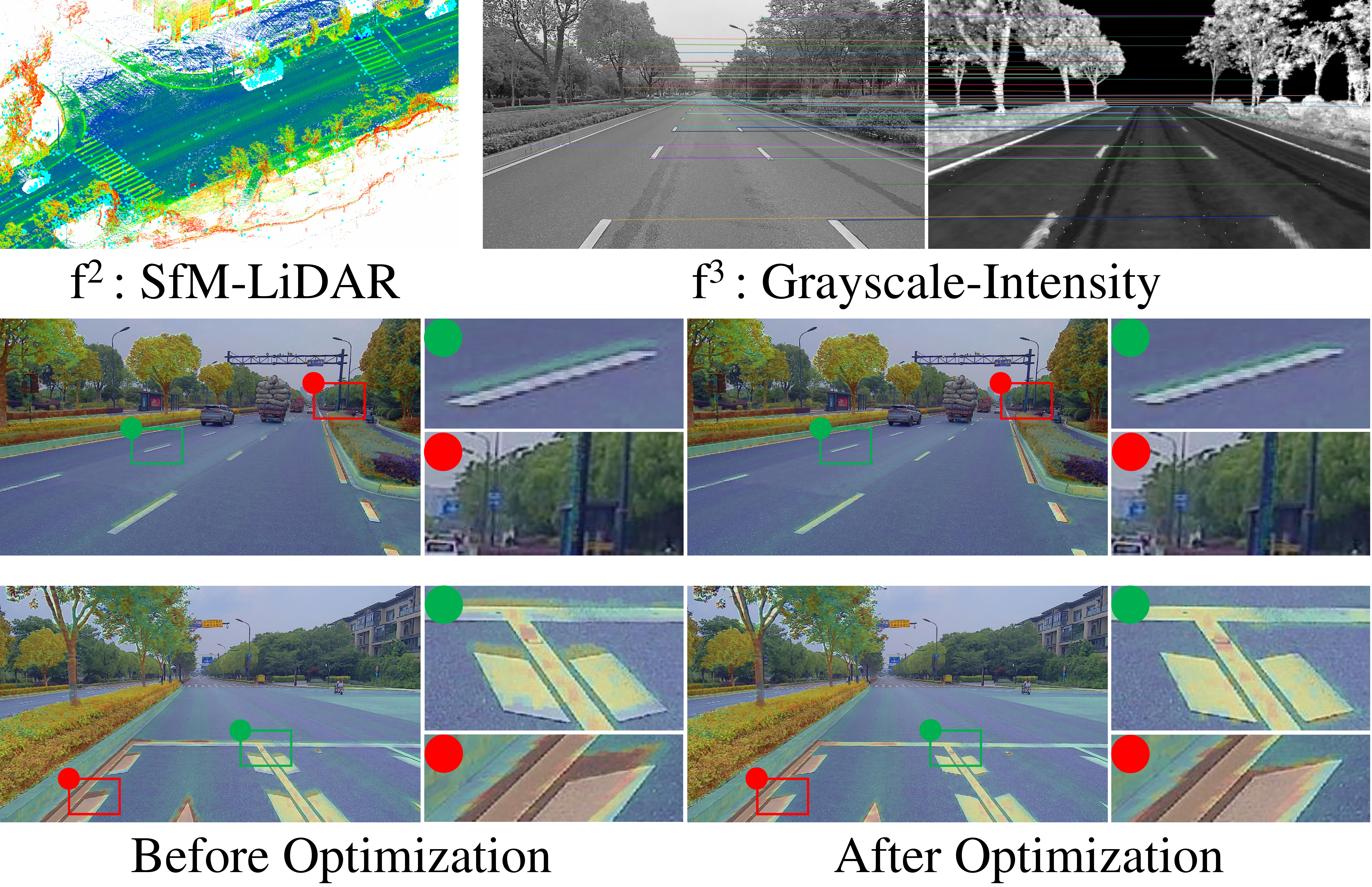}
\caption{Factors used in our cross-modal pose optimization framework, and we visualize LiDAR-camera alignment quality through an alpha-blending of the colorized intensity map onto its corresponding camera frame. {We refer readers to our supplementary video for the data alignment quality of our LiDAR map and multiple camera frames.}}
\label{fig:sfm}
\end{figure}

We present the results of the ablation study in Tab.~\ref{tab:sfm} and Fig.~\ref{fig:sfm}. These results demonstrate that the classical SfM framework ($\mathbf{f}^1$) can jointly solve poses between camera frames to ensure single-modal coherency. However, the lack of correspondence to the LiDAR map $\mathcal{L}^\mathbb{G}$ undermines multi-modal coherency. Therefore, an effective approach within the SfM framework is to construct the paired relationship between Visual-SfM points and their corresponding LiDAR points, providing a strong depth prior for triangulation—especially for landmarks established with a narrow baseline (e.g., landmarks on the road or curb while vehicles drive straight). Moreover, photometric factors based on 2D input and ray-casted images can further enhance multi-modal coherency, as our experiments indicate. This is because pixel-level correspondences and gradients operating on grayscale and intensity images, similar to depth map contours used for texture mapping~\cite{zhou2014colormapopt}, directly utilize raw input channels from different sensors.

%
\begin{table}[t]
\centering
\fontsize{8pt}{9.6pt}\selectfont
\setlength{\tabcolsep}{5pt}
\caption{Quantitative metrics for pose estimation. We choose the reprojection error (in pixel) as a common SfM metric~\cite{scho2016sfm} to evaluate coherency between camera frames, and the NID-loss~\cite{pascoe2015robust} to evaluate coherency between camera frames and the LiDAR map.}
\begin{tabular}{l|cccc}
\toprule
Metric & Init. $\mathbf{T}_{\mathbb{C}_i}^\mathbb{G}$ & Opt. $\mathbf{f}^{1}$ & Opt. $\mathbf{f}^{1-2}$ & Opt. $\mathbf{f}^{1-3}$ \\
\midrule
Reproj. Err. $\downarrow$ &  7.959 & \cccf{1.342} & \ccct{1.344} & \cccs{1.343} \\
\midrule
NID-Loss  $\downarrow$   &  \ccct{5.32} & 10.52 & \cccs{4.47} & \cccf{4.10} \\
\bottomrule
\end{tabular}

\label{tab:sfm}
\end{table}
\section{Benchmark}

Based on our scanners, scanning-pattern, and pose optimization procedures discussed in Sec.~\ref{sec:dataset}, we have prepared grouped sequences for benchmarking NVS methods on multi-lane scenarios. First of all, we discuss on implementation details of our selected NVS methods in Sec.~\ref{sec:bench:methods}, and then share our protocols and metrics chosen for cross-lane NVS application in Sec.~\ref{sec:bench:design}.

\subsection{Benchmarking Methods}
\label{sec:bench:methods}

We select a range of methods, specifically those designed for autonomous driving datasets and based on either NeRF or 3DGS, as our benchmarking methods.
For all methods discussed below, we use the combination of LiDAR $\mathcal{L}^\mathbb{G}$ and SfM $\mathcal{P}^\mathbb{G}$ points to initialize the Gaussians. 
In order to eliminate the influence of dynamic objects, we filter out all the cars and pedestrians in point clouds and images based on semantic labels.
Since the original 3DGS and most of its extensions only support focal points at the absolute center, we rectify the focal point to the center of corresponding images during the post-processing of our dataset.

\textbf{3DGS~\cite{Kerbl20233dgs}:} 
We utilize the implementation from the official release to evaluate our dataset and employ the AdamW optimizer with a learning rate of \(10^{-3}\). The model is trained for 30,000 steps.

\textbf{GaussianPro~\cite{cheng2024gaussianpro}:} 
We train the models for 30,000 iterations across all scenes, adhering to the original training schedule and hyperparameters. The interval step for the progressive propagation strategy is set to 20, with propagation performed three times. 

\textbf{Scaffold-GS~\cite{Lu2024scaffoldgs}:} 
Both the appearance and feature dimensions in the MLP are set to 32, and voxel size is set to 0.005. We adjust the initial and hierarchy factors for anchor growing to 16 and 4, respectively. The model is trained for 30,000 steps.

\textbf{2DGS~\cite{Huang20242DGS}:} 
We keep most parameters consistent with the original implementation. For densification, we adjust the gradient threshold to \(3 \times 10^{-4}\) and set the final densification iterations to 13,000. The model is trained for 30,000 steps.

\textbf{Street Gaussians~\cite{yan2024street}:} To ensure a fair comparison with other methods, we omit the handling of dynamic objects from the original scene graph method. Additionally, we do not utilize the sky mask for an equitable comparison. The parameters remain consistent with those in the official implementation.

\textbf{PVG~\cite{chen2023pvg}:} 
We utilize the Adam optimizer and keep a comparable learning rate for most parameters, consistent with the original implementation. We adjust the gradient threshold to \(3 \times 10^{-4}\) and set the final densification iterations to 13,000. The maximum number of Gaussian spheres is configured to \(10^{6}\). {We omit multi-resolution downsampling, using images at their original resolution for training.}

\textbf{EmerNeRF~\cite{yang2023emernerf}:} 
We train EmerNeRF for 30,000 iterations using its original parameters. The flow branch and temporal interpolation are activated, with both the feature levels of the hash encoder for the static and dynamic branches set to 4.

\subsection{Experimental Protocols and Metrics}
\label{sec:bench:design}

\begin{figure}[t]
\includegraphics[width=\linewidth]{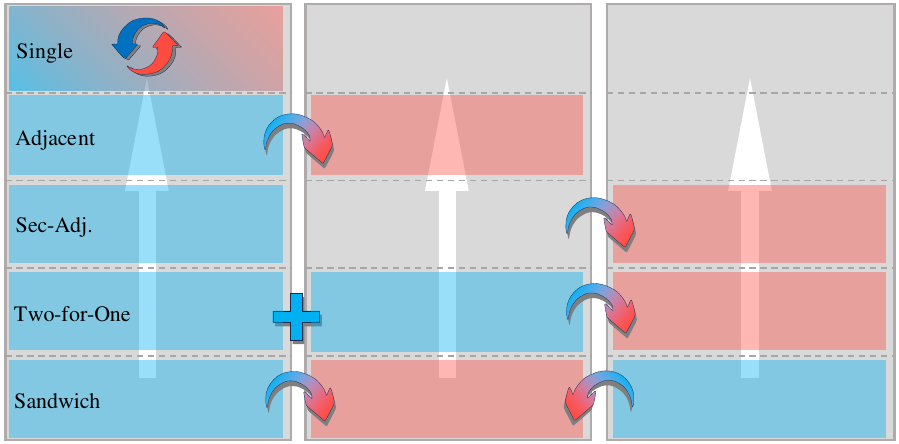}
\caption{{Five evaluation tracks using different combinations of lanes for training (colored in blue) and testing (colored in red).}}
\label{fig:expset}
\end{figure}

To perform a comprehensive benchmark of all the aforementioned methods on our proposed dataset, we meticulously group all sequences and organize the benchmarks across five different tracks for each method. Specifically, the tracks are categorized as follows: (1) Single lane regression, (2) Adjacent lane prediction, (3) Second-adjacent lane prediction, (4) Adjacent lane prediction (trained from two lanes), and (5) Sandwich lane prediction (trained from two side lanes). A figure illustrating the experimental protocols is provided in Fig.~\ref{fig:expset}. For each track, we uniformly sample 200 frames from training sequences for model learning and 25 frames from test sequences as the ground truth.

We adhere to the widely used metrics for evaluating the performance of NVS as outlined in \cite{li2024xld}, which includes Peak Signal-to-Noise Ratio (PSNR), Structural Similarity Index (SSIM), and Learned Perceptual Image Patch Similarity (LPIPS).

\subsection{Experimental Results and Notes}
\label{sec:bench:main}

\begin{table*}[t]
\caption{Quantitative results of different Gaussian reconstruction methods on our proposed Para-Lane dataset.}
\centering
{
\fontsize{8pt}{9.6pt}\selectfont
\setlength{\tabcolsep}{2.2pt}
\begin{tabular}{l|ccc|ccc|ccc|ccc|ccc}
\toprule
Method                  &   \hmerge{Single}     &  \hmerge{Adjacent}    &  \hmerge{Sec-Adj.}    &  \hmerge{Two-for-One} & \hmergl{Sandwich}       \\
Metrics                 &       \metrics        &      \metrics         &      \metrics         &      \metrics         &      \metrics           \\ 
\midrule
3DGS                    & \cccf{22.99} & \cccf{0.689} & \cccs{0.344} & \ccct{17.05} & \cccs{0.524} & \cccf{0.446} & \ccct{16.26} & \ccct{0.505} & \cccs{0.472} & \ccct{17.85} & \ccct{0.551} & \ccct{0.440} & 18.74 & \ccct{0.563} & \cccs{0.424}    \\
GaussianPro             & \ccct{22.93} & \cccs{0.687} & \cccf{0.343} & 17.01 & 0.521 & \cccf{0.446} & \cccs{16.29} & \ccct{0.505} & \cccs{0.472} & 17.83 & \ccct{0.551} & \cccs{0.439} & 18.66 & 0.562 & \cccs{0.424}   \\
Scaffold-GS             & \cccs{22.96} & \ccct{0.675} & 0.364 & \cccf{17.59} & \cccf{0.538} & \ccct{0.450} & \cccf{17.09} & \cccf{0.525} & \cccf{0.470} & \cccf{18.62} & \cccf{0.565} & \cccf{0.437} & \cccf{19.20} & \cccf{0.574} & \cccf{0.423}   \\
2DGS                    & 22.29 & 0.651 & 0.395 & 16.79 & \ccct{0.523} & 0.469 & 16.01 & \cccs{0.510} & 0.494 & 17.46 & 0.548 & 0.466 & \cccs{19.04} & \cccs{0.572} & 0.451   \\
Street Gaussians        & 22.56 & 0.643 & \ccct{0.353} & \cccs{17.50} & 0.510 & 0.456 & 16.16 & 0.496 & 0.480 & \cccs{17.87} & \cccs{0.555} & 0.453 & \ccct{18.91} & 0.561 & 0.443   \\
\bottomrule
\end{tabular}
}

\label{tab:eval}\vspace{-10pt}
\end{table*}

\begin{figure*}[t]
\includegraphics[width=\linewidth]{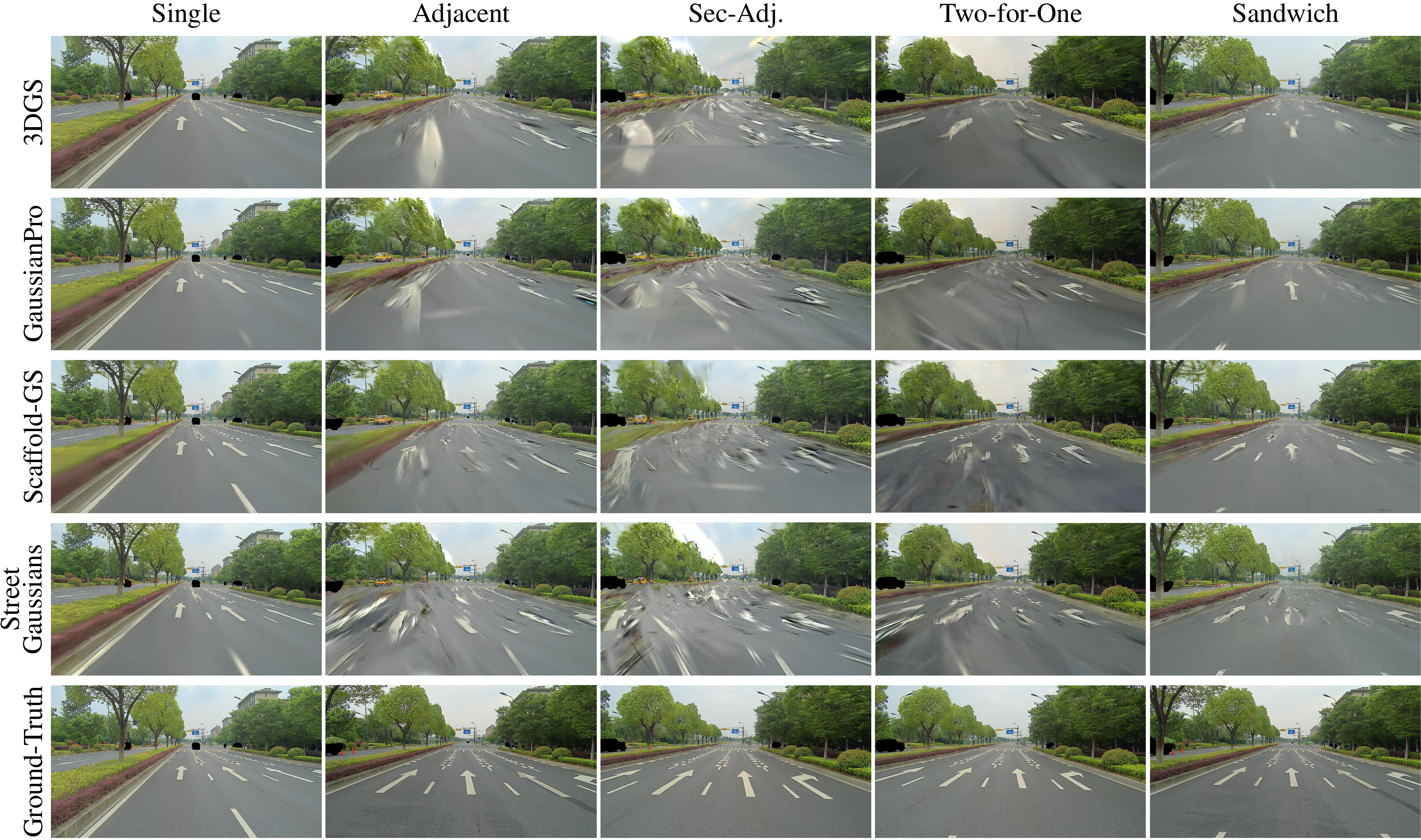}
\caption{Comparisons for NVS quality between different methods and designs, see our supplementary video for results on more sequences.}
\label{fig:eval}
\end{figure*}

We benchmark the methods described in Sec.~\ref{sec:bench:methods} according to the framework outlined in Sec.~\ref{sec:bench:design} to evaluate performance for cross-lane NVS. A series of experiments were conducted using an NVIDIA GeForce RTX 3090 24GB GPU. For quantitative metrics across all tested methods and different designs, refer to Tab.~\ref{tab:eval}. Throughout the experiments, we discovered several interesting insights:
\vspace{4pt}

\noindent \textbf{The quality of NVS is significantly affected by the view distribution of the training set.} 
From Tab.~\ref{tab:eval}, we found {\emph{exactly}} the same conclusion for all methods: the performance gradually decreases in the following sequence: Single $>$ Sandwich $>$ Two-for-One $>$ Adjacent $>$ Second-Adjacent. When the training and testing views are on the same trajectory, all methods achieve the best NVS results. However, when the testing viewpoint undergoes lateral shifts, the results are compromised to varying degrees. 
{Besides, although the number of images used for training is the same, our Sandwich {track}, which evenly distributes training views on both sides of the test views, consistently achieves superior rendering quality compared to the Two-for-One {track}, where training views are located only on one side of the testing views.}
This can be attributed to the fact that when training data is more evenly distributed and closer to the target render pose, the learned radiance functions are less likely to overfit to a specific viewpoint. 
Therefore, from an application perspective, to generate images of a target scene from arbitrary viewpoints, it is advisable to utilize multiple passes of data to reconstruct the target scene, thereby minimizing potential artifacts during novel view synthesis.
{Fig.~\ref{fig:eval} presents a visual comparison of novel view synthesis results under different designs, and our supplementary video provides additional examples. }


\noindent \textbf{Domain gap between synthetic~\cite{li2024xld} and real datasets.} Most of the methods tested here were also evaluated on the XLD dataset~\cite{li2024xld}, a synthetic cross-lane dataset; however, they generally did not perform as well on our dataset as they did on XLD. We attribute this difference to the domain gaps between synthetic and real-world data for several reasons. Firstly, synthetic data perfectly ensures the accuracy of all parameters, including extrinsic, intrinsic, and timestamps. However, in the real world, despite the optimizations discussed in Sec.~\ref{sec:dataset:slam}, there inevitably remains a gap between our final estimations and the ground truth values. Secondly, the sequences in a group were collected over different time intervals, leading to minor variations in brightness, water stain shapes, and other trivial factors among the cross-lane sequences. This also brings errors that require NVS approaches to handle. Finally, in real-world data, special attention must be given to dynamic objects. Unlike synthetic data, we cannot capture observations of the same dynamic object simultaneously across different locations. Although we used SAM to mask dynamic objects, the model's output is not always precise, and some noise remains.

\noindent \textbf{Scaffold-GS achieves the best NVS performance on our benchmark.} 
Unlike other methods, Scaffold-GS~\cite{Lu2024scaffoldgs} utilizes anchor points to distribute 3D Gaussians. Each anchor point is associated with multiple neural Gaussians. The attributes of these neural Gaussians—such as position, opacity, quaternion, scaling, and color—are determined by a multi-layer perceptron (MLP). The input features for the MLP include the relative poses from the anchor points to the camera views. Our cross-lane experimental results demonstrate that this method, which establishes correlations between view poses and rendering outcomes, is effective.

\subsection{Handling Dynamic Objects}

Besides the main experiment in Sec.~\ref{sec:bench:main} that reflects performance of methods in masked static scenes, we perform another experiment on those experiments capable of handling dynamic objects.

EmerNeRF~\cite{yang2023emernerf} and PVG~\cite{chen2023pvg} are two representative methods that contain procedures on handling dynamic objects in a self-supervised manner, we compare the two methods using {the Single track. We perform reconstruction through both methods with and without automatically labeled mask~\cite{Kirillov2023sam}}
The dataset used in this part includes 6 groups, which are part of the entire 25 groups dataset.

\begin{figure}[htbp]
\includegraphics[width=\linewidth]{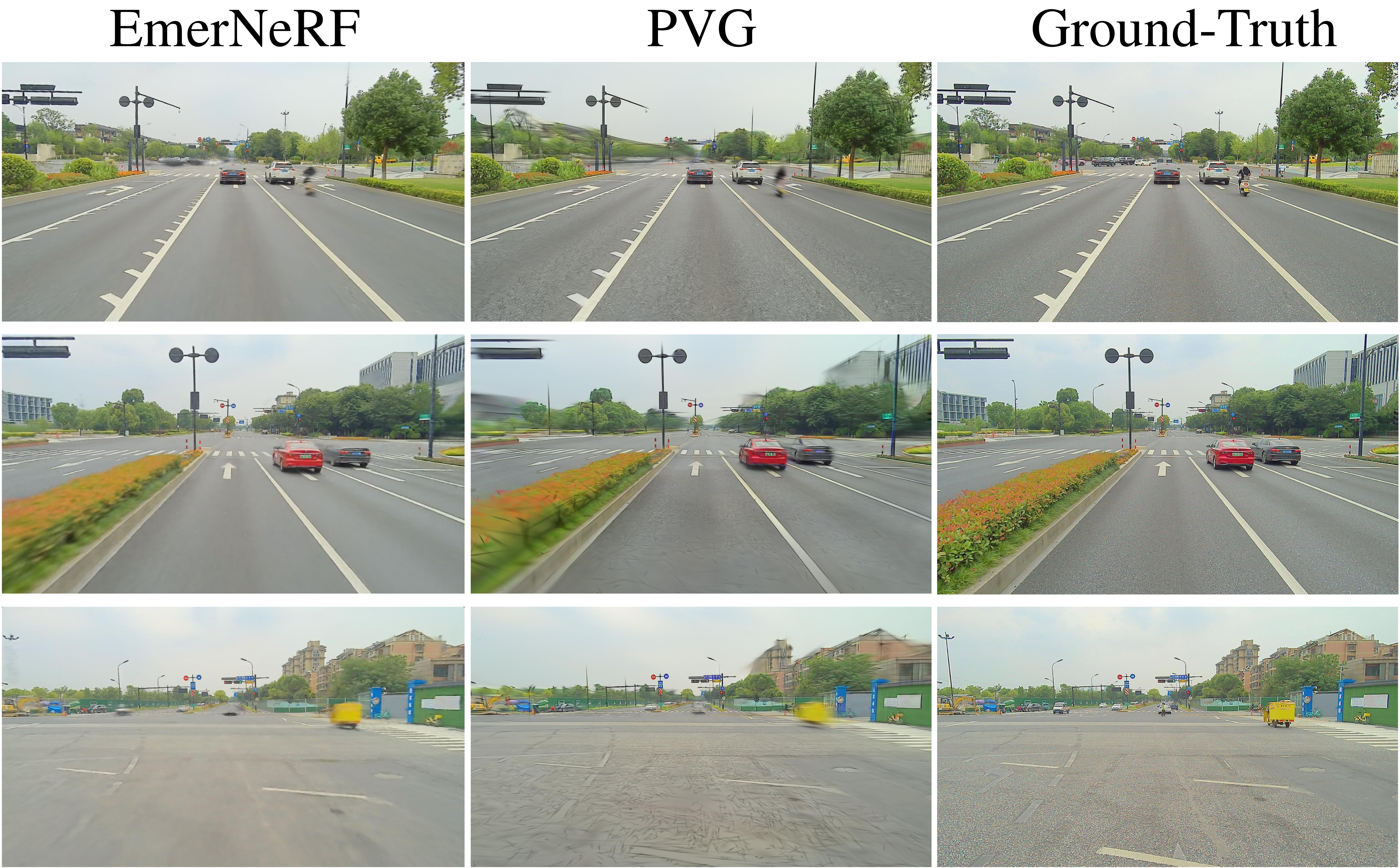}
\caption{Comparisons for NVS quality in Single lane test between EmerNeRF and PVG.}
\label{fig:pvg}
\end{figure}

\textbf{Results and analysis.} From Tab.~\ref{tab:eval_dynamic}, we can find that EmerNeRF achieves better PSNR and LPIPS scores. This is probably because of its novel and effective approach of static-dynamic decomposition. On the other hand, PVG excels in SSIM for both tests. This exceptional performance is likely to due to the adaptable design for Gaussian points. 
Qualitative results are shown in Fig.~\ref{fig:pvg}. We can find that the two methods are comparable for nearby scenes. However, PVG results are inferior for distant locations, such as buildings and trees. This is likely due to that PVG advocates to utilize the larger points for faraway scenes as described in its draft~\cite{chen2023pvg}, which results in inadequate expression of details and cause blur.

\begin{table}[htbp]
\centering
{
\fontsize{8pt}{9.6pt}\selectfont
\setlength{\tabcolsep}{5pt}
\begin{tabular}{l|ccc}
\toprule
Method                       &   \hmergl{Single}     \\
Metrics                      &       \metrics        \\ 
\midrule
EmerNeRF                     & \cccf{23.67} & 0.668 & \cccf{0.350} \\
PVG                          & 22.08 & \cccf{0.672} & 0.408 \\
\midrule
EmerNeRF (static only)       & \cccf{23.76} & 0.678 & \cccf{0.346} \\
PVG (static only)            & 22.77 & \cccf{0.684} & 0.368 \\
\bottomrule
\end{tabular}
}
\caption{Quantitative results on our proposed Para-Lane dataset with dynamic objects. We perform reconstruction with and without mask for ablation study.}
\label{tab:eval_dynamic}
\end{table}

\subsection{Limitations}

This section describes the limitations of our current dataset and benchmark. The dynamic object masks provided in the dataset are not manually labeled, so there are a small number of omissions and mislabeling. We also do not yet have 3D bounding box and tracking labels for all dynamic objects. The diversity of the dataset can also be enhanced by collecting more data in the future.
Given the fact that current works rarely support dynamic/static decomposition, we only tested EmerNeRF~\cite{yang2023emernerf} and PVG~\cite{chen2023pvg} in handling dynamic objects. A more comprehensive benchmark for dynamic scenes would be beneficial as more dynamic methods are developed in the future.
\section{Conclusion and Future Work}

In conclusion, we provide a two-stage framework that first registers multi-pass LiDAR frames to form a coherent map, and then registers camera frames to the LiDAR map for the multi-modal pose estimation. We use the provided method to produce reliable poses for multiple grouped parallel lane sequences, and test the performance of recent approaches for synthesizing novel views through longitudinal and lateral viewpoint shifts. 

In the future, we plan to expand our dataset with a variety of sequences and incorporate the latest approaches for thorough benchmarking. While we have successfully aligned fisheye camera frames to the LiDAR map, we have also identified limitations in recent works in jointly reconstructing with them. Handling such an issue presents an opportunity for the application of these methods in the industrial community, particularly for cost-effective labeling and mass production for closed-loop simulation scenarios.
We hope that the dataset we have released will facilitate future research in this area.

\section*{Acknowledgements}

We thank the reviewers for the valuable discussions and our colleagues for preparing the proposed dataset. This research was supported by the Zhejiang Provincial Natural Science Foundation of China under Grant No. LD24F030001.

{
    \small
    \bibliographystyle{ieeenat_fullname}
    \bibliography{main}
}

\end{document}